\documentclass[11pt,a4paper]{article}
\usepackage[hyperref]{emnlp2020}
\usepackage{times}
\usepackage{latexsym}

\usepackage{todonotes}
\usepackage{verbatim}
\usepackage{booktabs}
\usepackage{graphicx}
\usepackage{xcolor}
\usepackage[ruled,vlined]{algorithm2e}
\usepackage{multirow}
\usepackage{arydshln}

\newcommand{\bfsc}[1]{\textbf{\textsc{#1}}}

\usepackage{microtype}

\aclfinalcopy 

\newcommand\BibTeX{B\textsc{ib}\TeX}

\title{Narrative Interpolation for Generating and Understanding Stories}

\author{Su Wang \\
  Linguistics, UT Austin\\
  \footnotesize{\texttt{shrekwang@utexas.edu}} \\\And
  Greg Durrett \\
  Computer Science, UT Austin\\
  \footnotesize{\texttt{gdurrett@cs.utexas.edu}} \\\And
  Katrin Erk \\
  Linguistics, UT Austin\\
  \footnotesize{\texttt{katrin.erk@mail.utexas.edu}} \\}

\date{}

\begin{document}
\maketitle
\begin{abstract}
We propose a method for controlled narrative/story generation where we are able to guide the model to produce coherent narratives with user-specified target endings by interpolation: for example, we are told that \emph{Jim went hiking} and at the end \emph{Jim needed to be rescued}, and we want the model to incrementally generate steps along the way. The core of our method is an interpolation model based on GPT-2 which conditions on a previous sentence and a next sentence in a narrative and fills in the gap. Additionally, a reranker helps control for coherence of the generated text. With human evaluation, we show that ending-guided generation results in narratives which are coherent, faithful to the given ending guide, and require less manual effort on the part of the human guide writer than past approaches.
\end{abstract}

\section{Introduction}

Controllable storytelling \cite{Fan:18,See:19} is a challenging problem in natural language generation: a high quality story is logically consistent, topically coherent, and pragmatically reasonable \cite{Grice:75}. Previous approaches have guided this generation conditioned on topics \cite{Fan:18,Li:19} and ``storylines'' \cite{Li:13,Yao:18,Fan:19,Ammanabrolu:19}. The former requires minimal human inputs yet does not prevent ``on-topic-wandering'' --- a collection of events falling loosely under a given topic yet lacking a clear plot. The latter (more related to our work) can help produce high quality stories and offer minute content control, but needs the human guide to fit the granularity of the storyline appropriately and can be sensitive to the choices of keyphrases (e.g. Table~\ref{tab:case_paw_vs_interpol}).

\begin{table}[!t]
    \centering
    \scalebox{0.58}{
    \begin{tabular}{lll}
         \multicolumn{3}{l}{\textsc{beginning}: Jim went hiking alone at the state park.} \\
          \midrule
         He got lost on a trail.&He enjoyed the view.&He wandered off.\\
         He slipped and fell.&He saw many wildlife.&He lost his map.\\
         Jim broke his leg.&The camping was fun too.&Jim never returned home.\\
         \midrule
         Jim was rescued. & Jim enjoyed the trip. & His wife reported his missing. 
    \end{tabular}}
    \caption{Ending-guided story generation --- telling different stories in accordance with different endings. \textbf{Top row}: a beginning sentence for a story; \textbf{Bottom row}: three user-provided endings indicating different courses of events; \textbf{Mid rows}: interpolated bridging sentences (from model) with the same beginning and three respective endings. The leftmost story is written by a human, and the other two are generated by our model.}
    \label{tab:example_different_endings}
\end{table}

\begin{table}[!ht]
    \centering
    \scalebox{0.6}{
    \begin{tabular}{l}
         \textbf{Beginning}: \emph{Sarah cherished her favorite toy that her mother gave her.} \\
         \midrule
         \textcolor{red}{Storyline-guided model} (\textsc{paw}, \newcite{Yao:18}) \\
         \midrule
         \textbf{Storyline 1}: \texttt{played toy-toy disappeared-never found it-} \\
         \texttt{bought new toy} \\
         \textbf{Generated}: \emph{She loved to play toy with it every day. One day the toy} \\
         \emph{disappeared. Sarah searched all over the house and never found it.} \\
         \emph{She bought herself a new toy.} \vspace{0.5em}\\
         \hdashline \vspace{-0.5em}\\
         \textbf{Storyline 2}: \texttt{played-disappeared-lost-new} \\
         \textbf{Generated}: \emph{Sarah played in the courtyard. Her mother disappeared.} \\
         \emph{Sarah found herself lost in the woods. She bought a new house.} \\
         \midrule
         \textcolor{blue}{Ending-guided model} (\textsc{interpol}, ours) \\
         \midrule
         \textbf{Ending}: \emph{Her parents had to buy Sarah a new toy.} \\
         \textbf{Generated}: \emph{Sarah cherished her favorite toy that her mother gave her.} \\
         \emph{Sarah's mother forgot about the toy at school. Sarah was upset that} \\
         \emph{her mother had left the toy at school. Sarah was angry and cried.}
    \end{tabular}}
    \caption{Example: \textbf{storyline-guided} vs. \textbf{ending-guided} (ours) generation. The former is capable of generating high quality stories faithful to the conditioning storyline (storyline 1), but it can be sensitive to the choice of keyphrases, giving undesirable results with low-specificity storylines (storyline 2). Our method is not subject to this ``hyperparameter''. It only asks for a single ending sentence to produce good narratives.}
    \label{tab:case_paw_vs_interpol}
\end{table}

In light of these challenges, we propose a new method to model distributions of interpolated events in stories. We take as input a beginning sentence and an \emph{ending guide}, then generate a likely sequence of events in between to form a coherent story. Table~\ref{tab:example_different_endings} illustrates how different endings can lead to drastically different stories. We use a model that incrementally takes a pair of sentences and generates a sentence that might connect these two; this process starts with the beginning and ending of the story, but then conditions on our own generated intermediate sentences as they are produced (see Figure~\ref{fig:interpol} for an example). This generator is based on GPT-2 \cite{Radford:19}, which has seen successful application in past story generation work \cite{See:19}. We found one specific configuration of the generator to be particularly useful --- concatenating the right context before the left context and then generating the middle text. We additionally use a RoBERTa \cite{Liu:19} coherence ranking module to improve text quality and reject bad generations \cite{Ko:19,Guan:20}. 

Our results show that our method produces stories that are on-par or better than previous state-of-the-art guidance techniques \cite{Yao:18} in coherence, faithfulness, and overall preference according to human evaluation, despite the reduction in the amount of guidance inputs from the user.


\section{Related Work}

Past work has proposed several methods for controlled story generation \cite{Harrison:17,Martin:17,Mitchell:18,Lukin:18,Chandu:19}. Topic conditioned models \citep{Fan:18,Li:19} generate long-form stories from simple topic inputs, e.g., \emph{terrible scientific discovery}. 
Our work is most directly related to a line of techniques on storyline conditioning, which offer more fine-grained content control, and to which we compare below. \citet{Yao:18} in particular proposed a technique where a story can be fleshed out from a storyline consisted of a sequence of keyphrases (Table~\ref{tab:case_paw_vs_interpol}). Along this line, \citet{Ammanabrolu:19} traded the simplicity of keyphrase-based storylines for the stronger content control with event-based storylines. In another previous effort, \citet{Li:13} proposed a graph-based storyline-guided generation method. Our approach strikes a balance between these: our guidance is less fine-grained than full storylines but provides more guidance than just a general topic.

For generation quality, \citet{Ko:19} applied reranking to combat issues such as repetitiveness, contradition, etc. More recently \citet{Guan:20} added pre-trained LMs into the equation which improve text quality further in human evaluation. Both approaches use synthetically generated negative examples to target particular kinds of generation errors; we employ a similar kind of reranker to address key error types in our setting.
\section{Task and Data}

\paragraph{Task.} With the beginning sentence $b$ and a user-specified ending sentence $e$, we want to construct a complete story with pre-specified length (say, $k$ sentences). i.e. a function $f:(b, e; k) \rightarrow S_k$ where $S_k = \{b, s_1, \dots, s_{k-2}, e\}$ is a coherent story with $k-2$ sentences $\{s_1, \dots, s_{k-2}\}$ interpolated. Examples are presented in Table~\ref{tab:example_different_endings}.


\paragraph{Data.} Following Yao et al.~\shortcite{Yao:18}, which we directly compare to, we use the ROCStories short story dataset \cite{Mostafazadeh:17} --- a collection of 98,162 human-written 5-sentence English stories. The dataset has a vocabulary of 33,215 words, and an average sentence length of 10. We apply a 0.9/0.05/0.05 train/dev/test split.\footnote{This is our own random split of the data, since the split of \citet{Yao:18} was not available at the time of this writing. We will provide our datasets with the final version of this work for reproducibility.}

\section{Modeling}

We propose a generation-by-interpolation storytelling model: given the beginning of the story, and an ending as the guide for \emph{where the story should be going}, the model interpolates bridging events in between to construct a complete story at user-specified length. 

\begin{figure*}[!ht]
    \centering
    \includegraphics[width=130mm]{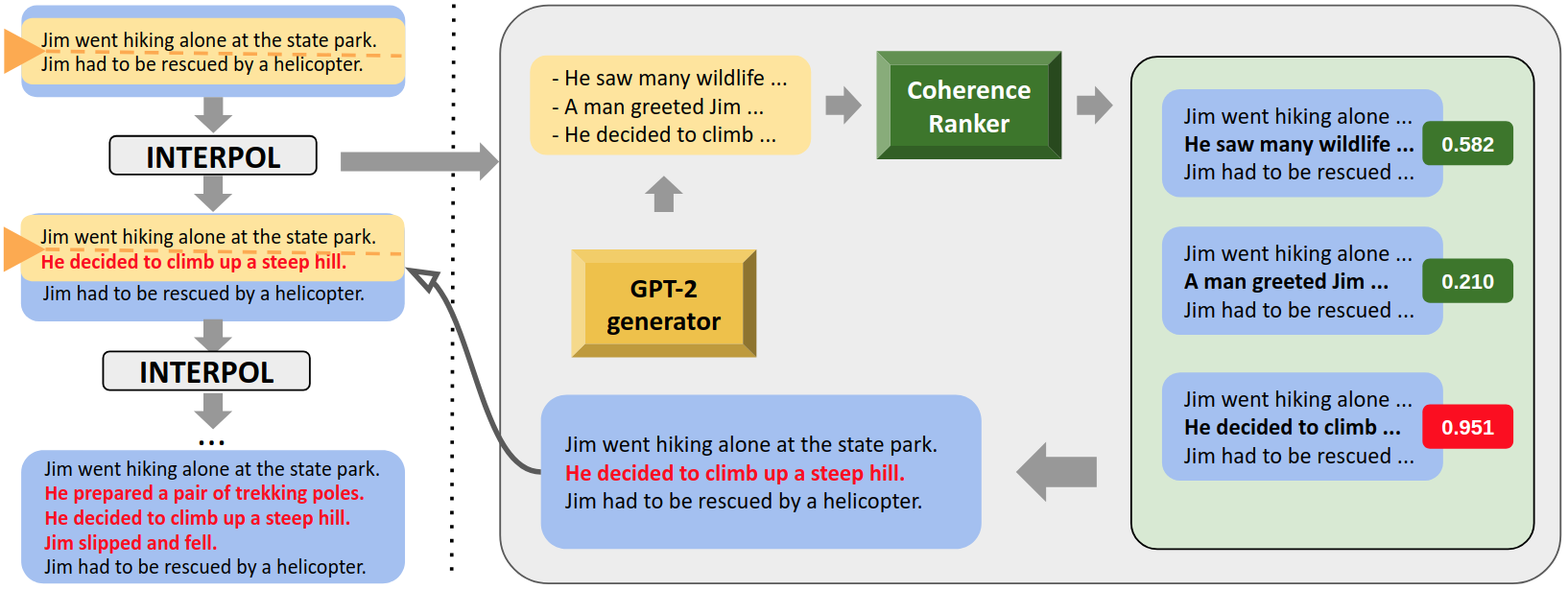}
    \caption{One step of our iterative narrative generation by \textsc{interpol}. \textbf{Left}: high-level flow of the interpolation procedure --- for each insertion point, produce an interpolation sentence. \textbf{Right}: detailed view of \textsc{interpol}. First, with a selected left-right context pair, the text generator proposes a list of interpolation candidates. Then the coherence ranker picks out the \textcolor{red}{\textbf{globally best candidate}} in the context of the story-in-construction. 
    NB: the order of interpolation is ``bisectional'': for a 5-sentence story, taking $s_1,s_5$ we generate $s_3$, then taking $s_1,s_3$ we generate $s_2$; finally generating $s_4$ given $s_3,s_5$.}
    \label{fig:interpol}
    \vspace{-0.2in}
\end{figure*}


Let $S_{init}=\{b,e\}$ be the initial input, where $b$ and $e$ are a story beginning and ending. We want to generate a $k$-sentence story $S_k$ with Eq. (1). The procedure runs iteratively as follows: given a story-in-construction $S$ and a desired story length $k$, randomly sample\footnote{We compared this approach to purely sequential interpolation, i.e. starting with $s_1,s_5$, generating $s_2$, then given $s_1,s_2,s_5$ and generate $s_3$ and so on. Measuring wordpiece perplexity we observed no statistically significant difference.} a possible insertion point (where both left and right context exist), then have a generation model propose $m$ viable interpolation candidates (with only the immediate left and right context as the input). A coherence ranker then evaluates each candidate in the context of the story by assigning a score based on the story-in-construction $S$ after insertion for each candidate. The ranker returns a best candidate $s$ to update the current story-in-construction. The procedure is described in Appendix D, and terminates when $S$ reaches length $k$. We call this model \bfsc{interpol} (\emph{interpolation}), which is shown in Figure~\ref{fig:interpol}.

For the \textbf{Text Generator}, we use OpenAI GPT-2 \cite{Radford:19} as our text generation model. We feed the model a pair of sentences to between which it should generate an interpolated sentence, and sample that interpolated sentence one word at a time. The inputs are the concatenation $[r;l]$, where the \emph{future} sentence appears first. This allows the left context to be closer to its direct continuation, localizing information in a way that has proven useful to past neural models \cite{Sutskever:14},while still conditioning on the completion.

The \textbf{Coherence Ranker} is implemented with pre-trained RoBERTa \cite{Liu:19}, and is based on the coherence evaluation techniques from prior work \cite{Ko:19,Moon:19,Guan:20}: the input $S$ is a story segment, as a list of tokens, and the output is a softmax over a binary label of whether that story is coherent or incoherent, trained with cross-entropy loss. During training, the model is supervised with 0/1 coherence labels, and during inference, the probability it produces for the positive class is treated as a real-valued \emph{coherence score}. The ranker is trained on a mix of positive coherence examples (real stories) and negative examples automatically generated to address a few chief classes of errors \cite{Guan:20}. For \emph{Repetition}, we randomly choose a sentence in a segment of a ROCStory (a number of consecutive sentences in the original) and duplicate it; for \emph{Irrelevant} we randomly sample a sentence from a different story to insert into a story segment; for \emph{Out-of-order} we randomly shuffle a subset of the sentences in a story segment.


\section{Experiments and Analysis}


\paragraph{Metrics} As \citet{See:19} state and show, human evaluation is currently the only reliable way to assess text quality in story generation. Adapting from related work \cite{Yao:18,Guan:20}, we apply three metrics for story quality: \textbf{Coherence}: whether a story is logically consistent or describes a possible scenario; \textbf{Faithfulness}: whether a story is faithful to a given story beginning and ending; \textbf{Overall Preference}: whether the human judge like a sequence of sentences \emph{as a story}. For concrete examples, see Appendix A.

\paragraph{Crowdsourced data for evaluation input.} To create (beginning, ending) pairs for evaluation time, we randomly sampled stories from the ROCStories test set and extracted the first and last sentences. To evaluate \emph{Faithfulness} when the ending changes with the same beginning, and to test model generalizability, we asked workers to write alternative endings for given (beginning, ending) pairs. Based on the three metrics above we elicit three alternative endings. For details, see Appendix B.

\paragraph{Baselines.} For empirical validation, we introduce baselines and compare to a model from past work \cite{Yao:18}. For the two baselines, the first is the same GPT-2, but without the right context sentence --- rather than interpolation, it only fine-tunes to generate the next sentence in a story, call it \textbf{\textsc{LeftToRight}}. The second evaluates the value of coherence ranking --- we use the interpolation GPT-2 model but generate a single candidate rather than select the best candidate out of multiple proposals, call it \textbf{\textsc{NoRanking}}. Finally we compare with a reformulated variant of Yao et al.~\shortcite{Yao:18}'s \emph{Plan-and-Write} (\textbf{\textsc{paw}}), based on their ``static'' variant.\footnote{The main difference is the original model also generates storylines, which can often be difficult to interprete (e.g. \texttt{work-fix-called-found-day}), whereas our reformulation uses human-written ones.} The model generates complete stories by conditioning on storylines as the context, where a storyline is a chain of keyphrases (see Table~\ref{tab:case_paw_vs_interpol}).

Following Yao et al.~\shortcite{Yao:18}, we use the RAKE algorithm to extract keyphrases from individual sentences to construct (storyline, story) training pairs. The first sentence of a story segment (as the context sentence) is prepended with a storyline extracted from the gold subsequent sentences in the segment, with type tokens to differentiate the storyline and the context sentence. To ensure strong comparability with our ending-guide condition, we took the three endings prepared above, then asked workers to write storylines with these as the last keyphrases. Examples in Appendix C.



\paragraph{Is the ending guide informative?} Our first question is whether the ending guide provides useful context to the generation model. We evaluate this on ROCStories by testing our models' ability to generate sentences 2-4 given either just sentence 1 (\textsc{LeftToRight}) or sentences 1 and 5 (\textsc{NoRanking}). In both the single-sentence generation condition and the full-story generation condition,
we compute the average perplexity of each wordpiece generation decision. Table~\ref{tab:bn_vs_interpol} summarizes the results, confirming the informativeness of the ending guide.

\begin{table}[!ht]
    \centering
    \scalebox{0.7}{
    \begin{tabular}{lcc|cc}
         \multirow{2}{*}{Model} & \multicolumn{2}{c|}{\emph{Single-sentence}} & \multicolumn{2}{c}{\emph{Full-story}} \\
         \cmidrule(lr){2-5}
         & \textsc{l2r} & \textsc{nr} & \textsc{l2r} & \textsc{nr} \\
         \midrule
         Perplexity & 8.90 & \textbf{6.76} & 9.93 & \textbf{7.53} \\
    \end{tabular}}
    \caption{\textsc{LeftToRight} (\textsc{l2r}) vs. \textsc{NoRanking} (\textsc{nr}), showing the informativeness of conditioning on the story ending (by wordpiece perplexity).}
    \label{tab:bn_vs_interpol}
\end{table}

\paragraph{Is coherence ranking helpful?} Next, we compare \textsc{NoRanking} (GPT-2 generator only) with \textsc{interpol} (GPT-2 generation + RoBERTa coherence ranking) as an ablation to gauge the value of coherence ranking. As human evaluation data we sampled 10 (beginning, ending) pairs from ROCStories test set,
and for each we crowdsourced 2 alternative endings, resulting in 30 entries in total. For each entry, both models generated a story which we presented to three workers. Table \ref{tab:NoRanking_vs_ranking} summarizes the results. The results demonstrate the effectiveness of coherence ranking: \textsc{interpol} is significanlty preferred on all three axes, indicating that the stories are overall higher-quality.

\begin{table}[!ht]
    \centering
    \scalebox{0.7}{
    \begin{tabular}{lcc|cc}
         & \textsc{nr} & \textsc{interpol} & &  \\
         & (-ranking) & (+ranking) & Both & Both \\
         & better & better & Good & Bad \\
         \midrule
         Coherence & 0.033 & \textbf{0.611} & 0.089 & 0.267 \\
         Preference & 0.078 & \textbf{0.589} & 0.044 & 0.289 \\
         \midrule
         & \textsc{nr} & \textsc{interpol} & & \\
         \cmidrule(lr){1-3}
         Faithfulness & 0.278 & \textbf{0.834} & \\
    \end{tabular}}
    \caption{Human eval on coherence ranking: \textsc{NoRanking} (\textsc{nr}) vs. \textsc{interpol}, showing humans greatly prefer the stories from the full \textsc{interpol} model.}
    \label{tab:NoRanking_vs_ranking}
\end{table}

\paragraph{Storyline-guide or ending-guide?} Finally, we compare \textsc{interpol} against \textsc{paw} on 90 sampled inputs where each is provided with its respective control sources (final sentences for \textsc{interpol} and storylines for \textsc{paw}). We demonstrate our method generates stories of equal or often better quality, summarized in Table \ref{tab:paw_vs_interpol}. In addition, as exemplified in Table~\ref{tab:case_paw_vs_interpol}, our guide inputs are more robust and easier to configure.

\begin{table}[!ht]
    \centering
    \scalebox{0.7}{
    \begin{tabular}{lcc|cc}
         & \textsc{paw} & \textsc{interpol} & Both & Both \\
         & better & better & Good & Bad \\
         \midrule
         Coherence & 0.178 & \textbf{0.444} & 0.233 & 0.144 \\
         Preference & 0.156 & \textbf{0.507} & 0.167 & 0.170 \\
         \midrule
         & \textsc{paw} & \textsc{interpol} & & \\
         \cmidrule(lr){1-3}
         Faithfulness & 0.333 & \textbf{0.744} & & 
    \end{tabular}}
    \caption{Comparison in Coherence, Faithfulness, and Overall Preference by human judgment: \textsc{paw} vs. \textsc{interpol}. Humans prefer stories generated by \textsc{interpol} across all axes, suggesting that ending conditioning is more effective than storyline conditioning.}
    \label{tab:paw_vs_interpol}
\end{table}


\section{Conclusion}

We propose a simple, intuitive yet effective method for storytelling, as an attempt to contribute to better narrative understanding. Our models leverage modern pre-trained language models, and with carefully constructed human evaluation we show 1) the models generate more coherent stories than previous work, and 2) the content of the stories can be easily and flexibly controlled by human user, who also prefer our generation \emph{as stories}. 

Our findings cast light on certain future directions: large-scale language models can manage commonsense inference only to a small extent, indicating the necessity of introducing human-provided knowledge. Further, beyond a general level of characterization of good stories (by coherence, faithfulness to human guide, etc), we can ask what specific organization of events and their participants constitutes good stories or narratives.

\bibliographystyle{acl_natbib}
\bibliography{anthology,emnlp2020}

\appendix

\end{document}